\documentclass[11pt,a4paper]{article}
\usepackage[hyperref]{acl2021}
\usepackage{times}
\usepackage{latexsym}
\usepackage{mathrsfs}
\usepackage{CJKutf8}
\usepackage{url}
\usepackage{makecell}
\usepackage{graphicx}  
\usepackage{mathrsfs}
\usepackage{amsmath}
\usepackage{bm}
\usepackage{multirow}
\usepackage{subfigure}
\usepackage{makecell}
\usepackage{array}
\usepackage{booktabs}
\usepackage{algorithm}
\usepackage{amssymb}
\usepackage{algpseudocode}

\usepackage{microtype}

\aclfinalcopy 

\title{A Joint Model for Dropped Pronoun Recovery and Conversational Discourse Parsing in Chinese Conversational Speech}

\author{Jingxuan Yang$^1$, Kerui Xu$^1$, Jun Xu$^{2,3*}$, Si Li$^{1}$, Sheng Gao$^1$, Jun Guo$^1$, \\
\textbf{Nianwen Xue$^4$ and Ji-Rong Wen$^{2,3}$} \\
$^1$School of Artificial Intelligence, Beijing University of Posts and Telecommunications \\
$^2$Gaoling School of Artificial Intelligence, Renmin University of China\\
$^3$Beijing Key Laboratory of Big Data Management and Analysis Methods\\
$^4$Department of Computer Science, Brandeis University \\
{\tt \{yjx, xukerui, lisi, gaosheng, guojun\}@bupt.edu.cn} \\
{\tt junxu@ruc.edu.cn, jirong.wen@gmail.com, xuen@brandeis.edu} 
  }

\date{}

\begin{document}
\begin{CJK*}{UTF8}{gbsn}
\maketitle

\begin{abstract}
In this paper, we present a neural model for joint dropped pronoun recovery (DPR) and conversational discourse parsing (CDP) in Chinese conversational speech. We show that DPR and CDP are closely related, and a joint model benefits both tasks. We refer to our model as DiscProReco, and it first encodes the tokens in each utterance in a conversation with a directed Graph Convolutional Network (GCN). The token states for an utterance are then aggregated to produce a single state for each utterance. The utterance states are then fed into a biaffine classifier to construct a conversational discourse graph. A second (multi-relational) GCN is then applied to the utterance states to produce a discourse relation-augmented representation for the utterances, which are then fused together with token states in each utterance as input to a dropped pronoun recovery layer. The joint model is trained and evaluated on a new Structure Parsing-enhanced Dropped Pronoun Recovery (SPDPR) dataset that we annotated with both two types of information. Experimental results on the SPDPR dataset and other benchmarks show that DiscProReco significantly outperforms the state-of-the-art baselines of both tasks.\let\thefootnote\relax\footnotetext{$^*$ Corresponding author}   
\end{abstract}

\section{Introduction}
Pronouns are often dropped in Chinese conversations as the identity of the pronoun can be inferred from the context~\citep{Kim2000Subject, Yang:15} without causing the sentence to be incomprehensible. The task of dropped pronoun recovery (DPR) aims to locate the position of the dropped pronoun and identify its type. Conversational discourse parsing (CDP) is another important task that aims to analyze the discourse relations among utterances in a conversation, and plays a vital role in understanding multi-turn conversations.

\begin{figure}[!t]
	\centering
	\includegraphics[width=7.0cm, height=7.5cm]{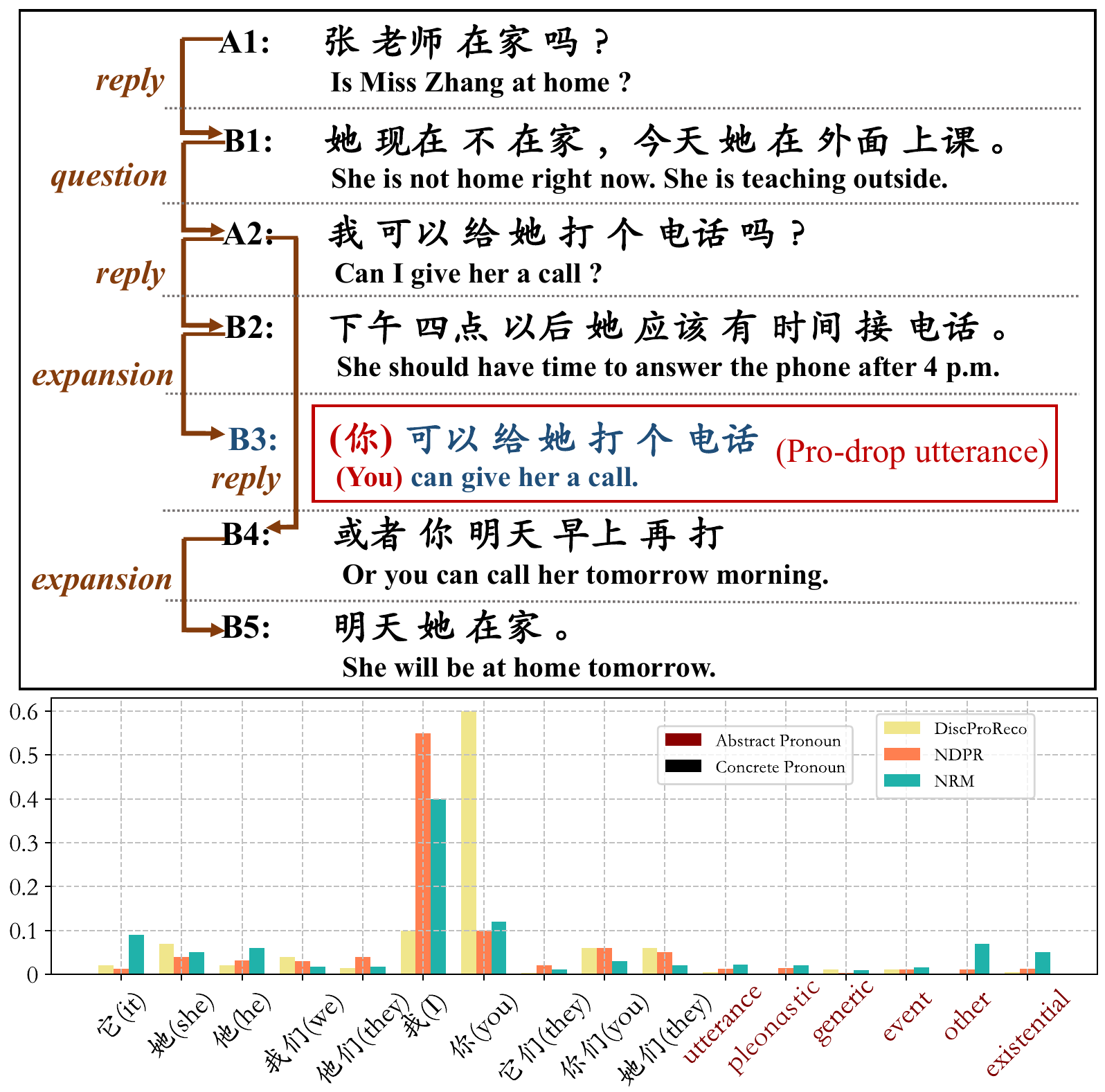}
	\caption{\textbf{Top:} A conversation snippet in which the dropped pronoun is shown in bracket. \textbf{Bottom:} Pronoun recovery results by two baselines and the proposed DiscProReco. Baselines which ignore the relation ``(B3 expands B2) replies A2'' mistakenly recover the dropped pronoun 你(you) as 我(I) since the utterance $\mathrm{B}_3$ is considered semantically similar to $\mathrm{A}_2$.}
	\label{mode_example}
\end{figure}

Existing work regards DPR and CDP as two independent tasks and tackles them separately. As an early attempt of DPR,~\citet{Yang:15} employ a Maximum Entropy classifier to predict the position and type of dropped pronouns.~\citet{zhang:neural} and~\citet{Yang2019NDPR} attempt to recover the dropped pronouns by modeling the referents with deep neural networks. More recently,~\citet{Yang2020} attempt to jointly predict all dropped pronouns in a conversation snippet by modeling dependencies between pronouns with general conditional random fields. A major shortcoming of these DPR methods is that they overlook the discourse relation \textit{(e.g., reply, question)} between conversational utterances when exploiting the context of the dropped pronoun. At the same time, previous CDP methods~\citep{li2014recursive, afantenos2015discourse, shi2018a} first predict the relation for each utterance pair and then construct the discourse structure for the conversation with a decoding algorithm. The effectiveness of these methods are compromised since the utterances might be incomplete when they have dropped pronouns.

To overcome these shortcomings, we propose a novel neural model called DiscProReco to perform DPR and CDP jointly. Figure~\ref{mode_example} is a Chinese conversation snippet between two speakers A and B that illustrates the advantages of such a joint approach. In this example, a pronoun ``你~ (you)'' is dropped in utterance $\mathrm{B}_3$. It is critical for the DPR model to know that both utterances $\mathrm{B}_2$ and $\mathrm{B}_3$ are in \emph{reply} to the utterance $\mathrm{A}_2$, when recovering this dropped pronoun. Methods which ignore the structure (``(B3 expands B2) replies A2'') will more likely consider the utterance $\mathrm{B}_3$ to be semantically similar to $\mathrm{A}_2$, and wrongly recover the pronoun as ``我~ (I)''.

Given a pro-drop utterance and its context, DiscProReco parses the discourse structure of the conversation and recovers the dropped pronouns in the utterance in four steps: (i) Each utterance is parsed into its dependency structure and fed into a directed GCN to output the syntactic token states. The utterance state is then obtained by aggregating the token states in the utterance. (ii) The utterance states of a conversation are fed into a bi-affine classifier to predict the discourse relation between each utterance pair and the discourse structure of the conversation is constructed. (iii) Taking the discourse structure as input, another (multi-relational) GCN updates the utterance states and fuses them into the token states for each utterance to produce discourse-aware token representations. (iv) Based on the discourse structure-aware context representation, a pronoun recovery module is designed to recover the dropped pronouns in the utterances. When training this model, all components are jointly optimized by parameter sharing so that CDP and DPR can benefit each other.
As there is no public dataset annotated with both dropped pronouns and conversational discourse structures, we also construct \underline{S}tructure \underline{P}arsing-enhanced \underline{D}ropped \underline{P}ronoun \underline{R}ecovery (SPDPR) corpus, which is the first corpus annotated with both types of information. Experimental results show that DiscProReco outperforms all baselines of CDP and DPR.

\noindent \textbf{Contributions: }This work makes the following contributions:
(i) We propose a unified framework DiscProReco to jointly perform CDP and DPR, and show that these two tasks can benefit each other. 
(ii) We construct a new large-scale dataset SPDPR (Section~\ref{dataset_construct}) which supports fair comparison across different methods and facilitates future research on both DPR and CDP. 
(iii) We present experimental results which show that DiscProReco with its joint learning mechanism realizes knowledge sharing between its CDP and DPR components and results in improvements for both tasks (Section~\ref{experiment}). 
The code and SPDPR dataset is available at \texttt{\url{https://github.com/ningningyang/DiscProReco}}.

\begin{figure*}
	\centering
	\includegraphics[width=14.4cm, height=8cm]{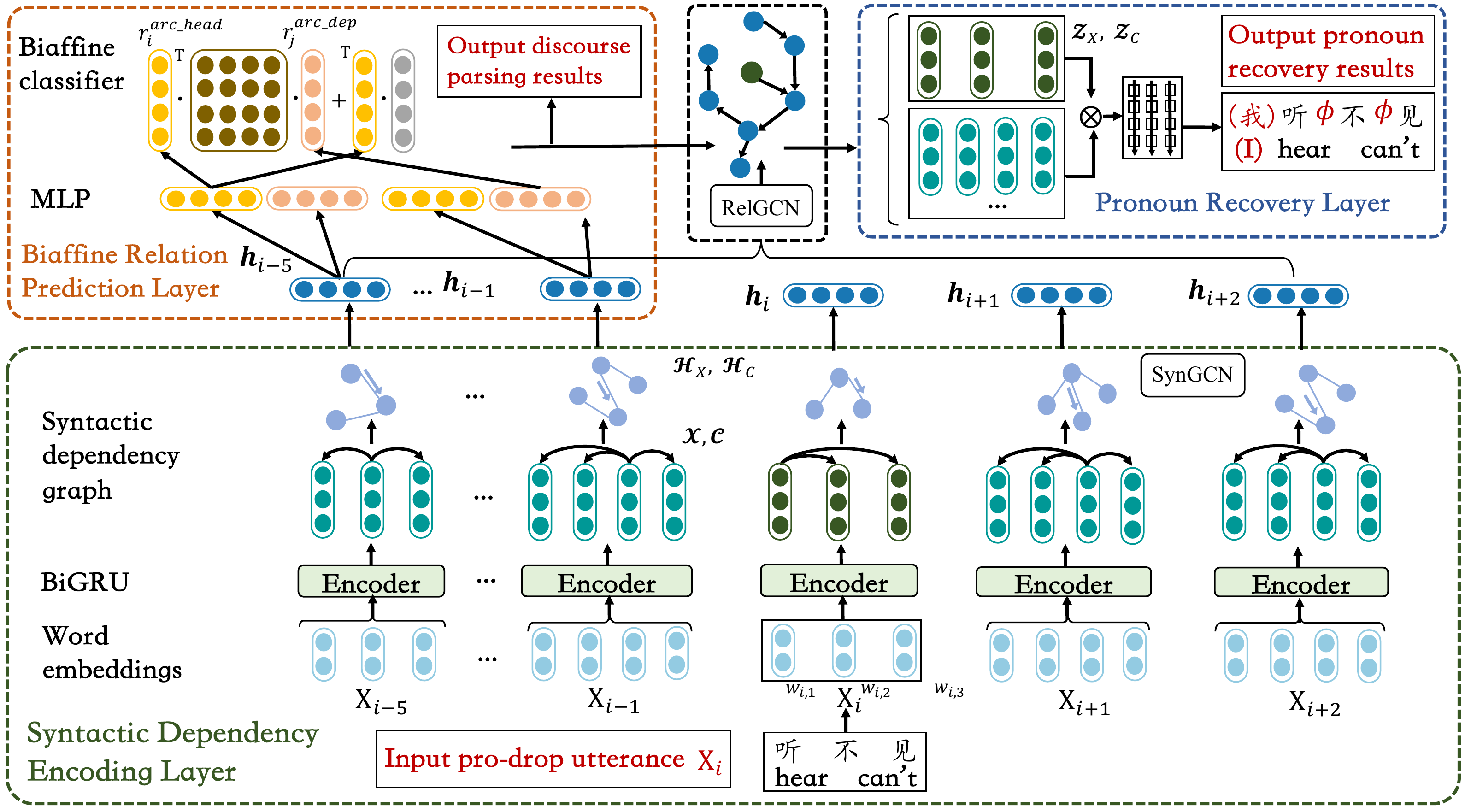}
	\caption{Overview of DiscProReco, which explores conversational discourse structures to learn effective referent representations that are used to recover dropped pronouns. DiscProReco consists of four components, and the details are introduced in Section~\ref{model_description}.
	}
	\label{framework}
\end{figure*}

\section{Problem Formulation}
We first introduce the problem formulation of these two tasks. Following the practices in~\citep{Yang:15, Yang2019NDPR, Yang2020}, we formulate DPR as a sequence labeling problem. DPR aims to recover the dropped pronouns in an utterance by assigning one of 17 labels to each token that indicates the type of pronoun that is dropped before the token~\citep{Yang:15}. CDP is the task of constructing the conversational discourse structure by predicting the discourse relation~\citep{xue2016annotating} among utterances. The discourse relations may characterize one utterance as agreeing with, responding to, or indicate understanding of another utterance in the conversational context. 

Let us denote an input pro-drop utterance of $n$ tokens as $\mathrm{X}=(w_1, w_2, \cdots, w_n)$, and its contextual utterances as $\mathrm{C}=(\mathrm{X}_1, \mathrm{X}_2, \cdots, \mathrm{X}_m)$ where the $i$-th contextual utterance $\mathrm{X}_i$ is a sequence of $l_i$ tokens: $\mathrm{X}_i=(w_{i,1},\cdots, w_{i,l_i})$. Our task aims to (1) model the distribution $\mathrm{P(X_j|X_i,C)}$ to predict the relation between each pair of utterances (i.e., $(\mathrm{X}_i, \mathrm{X}_j)$) for CDP, and (2) model $\hat{\mathrm{Y}}=\arg\max_{\mathrm{Y}} \mathrm{P(Y|X,C)}$ to predict the recovered pronoun sequence $\hat{\mathrm{Y}}$ for the input utterance $\mathrm{X}$. Each element of $\hat{\mathrm{Y}}$ is chosen from one of the $\mathrm{T}$ possible labels from $\mathcal{Y}=\{y_1, \cdots, y_{\mathrm{T}-1}\}\cup \{\mathrm{None}\}$ to indicate whether a pronoun is dropped before the corresponding token in utterance $\mathrm{X}$ and the type of the dropped pronoun. The label ``None'' means no pronoun is dropped before this token.

\section{The DiscProReco Framework}
\label{model_description}

\subsection{Model Overview}
The architecture of DiscProReco is illustrated in Figure~\ref{framework}. Given a pro-drop utterance $\mathrm{X}$ and its context $\mathrm{C}$, DiscProReco first represents tokens of these utterances as $d$-dimensional pre-trained word embeddings~\citep{DBLP:conf/acl/LiZHLLD18}, and then feed them into a BiGRU~\citep{chung2014empirical} network, to represent sequential token states $\boldsymbol{\mathcal{X}} \in \mathbb{R}^{n \times d}$ and $\boldsymbol{\mathcal{C}} \in \mathbb{R}^{m \times l_m \times d}$ as the concatenation of forward and backward hidden states outputted from BiGRU. The syntactic dependency encoding layer then revises the sequential token states by exploiting the syntactic dependencies between tokens in the same utterance using a directed GCN and generates utterance representations. After that, the biaffine relation prediction layer predicts the relation between each pair of utterances. The discourse structure then is constructed based on the utterance nodes and the predicted relations. The discourse structure encoding layer further encodes the inter-utterance discourse structures with a multi-relational GCN, and employs the discourse-based utterance representations to revise the syntactic token states. Finally, the pronoun recovery layer explores the referent semantics from the context $\mathrm{C}$ and predicts the dropped pronouns in each utterance.

\subsection{Syntactic Dependency Encoding Layer}
As the sequential token states overlook long-distance dependencies among tokens in a utterance, this layer takes in the sequential token states $\boldsymbol{\mathcal{X}}$ and $\boldsymbol{\mathcal{C}}$, and revises them as syntactic token states as $\boldsymbol{\mathcal{H}}_\mathrm{X}$ and $\boldsymbol{\mathcal{H}}_\mathrm{C}$ by exploring the syntactic dependencies between the tokens based on a directed GCN.

Specifically, for each input utterance in $\mathrm{X}$ and $\mathrm{C}$, we first extract syntactic dependencies between the tokens with Stanford's Stanza dependency parser~\citep{qi2020stanza}. Using the output of the dependency parser, we construct a syntactic dependency graph for each utterance in which the nodes represents the tokens and the edges correspond to the extracted syntactic dependencies between the tokens. 
Following the practices of ~\citep{marcheggiani2017encoding, Shikhar2018Dating}, three types of edges are defined in the graph. The node states are initialized by the sequential token states $\boldsymbol{\mathcal{X}}$ and $\boldsymbol{\mathcal{C}}$, and then message passing is performed over the constructed graph using the directed GCN~\cite{Kipf2016Semi}, referred to as \textbf{SynGCN}. The syntactic dependency representation of token $w_{i,n}$ after $(k+1)$-th GCN layer is defined as:
\[\resizebox{1\hsize}{!}{$
\label{dir_lab_gra}
\boldsymbol{h}_{w_{i,n}}^{k+1}=\mathrm{ReLU}\left(\sum_{u \in \mathcal{N}_{+}(w_{i,n})} g_e^{k} \cdot\left(\boldsymbol{\mathrm{W}}_e^{k} \boldsymbol{h}_{u}^{k}+\boldsymbol{b}_e^{k}\right)\right),$}
\]
where $\boldsymbol{\mathrm{W}}_e^{k} \in \mathbb{R}^{d \times d}$ and $\boldsymbol{b}_e^{k} \in \mathbb{R}^d$ are the edge-specific parameters, $\mathcal{N}_{+}(w_{i,n}) = \mathcal{N}(w_{i,n}) \cup \{w_{i,n}\}$ is the set of $w_{i,n}$'s neighbors including itself, and ReLU$(\cdot) = \max(0,\cdot)$ is the Rectified Linear Unit. $g_e^{k}$ is an edge-wise gating mechanism which incorporates the edge importance as:
\[
g_e^{k}=\sigma\left(\hat{\boldsymbol{w}}_e^{k} \boldsymbol{h}_{u}^{k}+\hat{b}_e^{k}\right),
\]
where $\hat{\boldsymbol{w}}_e^{k} \in \mathbb{R}^{1 \times d}$ and $\hat{b}_e^{k} \in \mathbb{R}$ are independent trainable parameters for each layer, and $\sigma (\cdot)$ is the sigmoid function. The revised syntactic token states $\boldsymbol{\mathcal{H}}_\mathrm{X}$ and $\boldsymbol{\mathcal{H}}_\mathrm{C}$ of the pro-drop utterance and context are outputted for subsequent discourse structure prediction and pronoun recovery.

\subsection{Biaffine Relation Prediction Layer}
For conversational discourse parsing, we jointly predict the arc $s_{i,j}^{(\operatorname{arc})}$ and relation $\mathbf{s}_{i,j}^{(\operatorname{rel})}$ between each pair of utterances utilizing the biaffine attention mechanism proposed in~\citep{Dozat2016}. Given the syntactic token states $\boldsymbol{\mathcal{H}}_\mathrm{X}$ and $\boldsymbol{\mathcal{H}}_\mathrm{C}$, we make an average aggregation on these token states of each utterance $\mathrm{X}_i$ to obtain the syntactic utterance representation $\boldsymbol{h}_{\mathrm{X}_i}$.

For a pair of utterances $(\mathrm{X}_i, \mathrm{X}_j)$ in the conversation snippet, we feed the representations of these two utterances into a biaffine function to predict the probability of an arc from $\mathrm{X}_i$ to $\mathrm{X}_j$ as:
\[\boldsymbol{r}_i^{(\operatorname{arc\_head})} = \mathrm{MLP}^{(\operatorname{arc\_head})} (\boldsymbol{h}_{\mathrm{X}_i}),\]
\[\boldsymbol{r}_j^{(\operatorname{arc\_dep})} = \mathrm{MLP}^{(\operatorname{arc\_dep})} (\boldsymbol{h}_{\mathrm{X}_j}),\]
\[\resizebox{1\hsize}{!}{$s_{i,j}^{(\operatorname{arc})} = \boldsymbol{r}_i^{(\operatorname{arc\_head})}\mathbf{U}^{(\operatorname{arc})}\boldsymbol{r}_j^{(\operatorname{arc\_dep})}+\boldsymbol{r}_i^{(\operatorname{arc\_head})^T}\boldsymbol{u}^{(\operatorname{arc})}$},\]
where MLP is the multi-layer perceptron that transforms the original utterance representation $\boldsymbol{h}_{\mathrm{X}_i}$ and $\boldsymbol{h}_{\mathrm{X}_j}$ into head or dependent-specific utterance states $\boldsymbol{r}_i^{(\operatorname{arc\_head})}$ and $\boldsymbol{r}_j^{(\operatorname{arc\_dep})}$. $\mathbf{U}^{(\operatorname{arc})}$ and $\mathbf{u}^{(\operatorname{arc})}$ are weight matrix and bias term used to determine the probability of a arc.

One distinctive characteristics of conversational discourse parsing is that the head of each dependent utterance must be chosen from the utterances before the dependent utterance. Thus we add an upper triangular mask operation on the results of arc prediction to regularize the predicted arc head:
\[
\mathbf{s}^{(\operatorname{arc})} = \mathrm{mask}(\mathbf{s}^{(\operatorname{arc})}).
\]

We minimize the cross-entropy of gold head-dependent pair of utterances as:
\[
\operatorname{loss_{arc}} = -\sum_{j=1}^{m} \delta(\mathrm{X}_{j} | \mathrm{X}_{i}, \mathrm{C}) \log (\operatorname{P_{arc}}(\mathrm{X}_{j} | \mathrm{X}_{i}, \mathrm{C})),\]
\[
\operatorname{P_{arc}}(\mathrm{X}_{j} | \mathrm{X}_{i}, \mathrm{C})=\operatorname{softmax}(\mathbf{s}_i^{( \operatorname{arc})}).
\]


After obtaining the predicted directed unlabeled arc between each utterance pair, we calculate the score distribution $\mathbf{s}_{i, j}^{\operatorname{(rel)}} \in \mathbb{R}^{k}$ of each arc $\mathrm{X}_i \rightarrow \mathrm{X}_j$, in which the $t$-th element indicates the score of the $t$-th relation as the arc label prediction function in~\citep{Dozat2016}. In the training phase, we also minimize the cross-entropy between gold relation labels and the predicted relations between utterances as:
\[\resizebox{1\hsize}{!}{$
\operatorname{loss_{rel}} = -\sum_{j=1}^{n} \delta(\mathrm{X}_{j} | \mathrm{X}_{i}, \mathrm{C}) \log (\operatorname{P_{rel}}(\mathrm{X}_{j} | \mathrm{X}_{i}, \mathrm{C})),$}
\]
\[
\operatorname{P_{rel}}(\mathrm{X}_{j} | \mathrm{X}_{i}, \mathrm{C})=\operatorname{softmax}(\mathbf{s}_{i, j}^{\operatorname{(rel)}}).
\]

\subsection{Discourse Structure Encoding Layer}
After the relations are predicted, we construct the discourse structure as a multi-relational graph in which each node indicates an utterance, and each edge represents the relation between a pair of utterances. In order to utilize the  discourse information in dropped pronoun recovery process, we first encode the discourse structure, and then utilize the discourse information-based utterance representations to improve token states which are used to model the pronoun referent.

Specifically, we apply a multiple relational GCN~\citep{vashishth2020composition}, referred to as \textbf{RelGCN}, over the graph to encode the discourse structure based utterance representations $\boldsymbol{\mathcal{R}}$ and utilize the updated representations to further revise syntactic token states $\boldsymbol{\mathcal{H}}_\mathrm{X}$ and $\boldsymbol{\mathcal{H}}_\mathrm{C}$ for outputting discourse structure based token states $\boldsymbol{\mathcal{Z}}_\mathrm{X}$ and $\boldsymbol{\mathcal{Z}}_\mathrm{C}$. The node states of the graph are initialized as the average aggregation of token states of corresponding utterances. The representation of utterance $\mathrm{X}_i$ in the $(k+1)$-th layer is updated by incorporating the discourse relation state $\boldsymbol{h}_{rel}^{k}$ as:
\[
\resizebox{1\hsize}{!}{$
\label{rel_gate_update}
\boldsymbol{r}_i^{k+1}=f\left(\sum_{(j, rel) \in \mathcal{N}(\mathrm{X}_i)}\operatorname{P_{rel}}(\mathrm{X}_{j}|\mathrm{X}_{i}, \mathrm{C})\boldsymbol{\mathrm{W}}_\mathrm{\lambda(rel)}^k \phi\left(\boldsymbol{r}_j^k, \boldsymbol{h}_{rel}^k\right)\right),$}
\]
where $\boldsymbol{r}_{j}^{k}$ and $\boldsymbol{h}_{rel}^{k}$ denote the updated representation of utterance $j$ and relation $rel$ after the $k$-th GCN layers, and $\boldsymbol{\mathrm{W}}_{\lambda(rel)}^{k} \in \mathbb{R}^{d \times d}$ is a relation-type specific parameter. Following the practice of ~\citep{vashishth2020composition}, we take the composition operator $\phi$ as multiplication in this work. Please note that we take in the label distribution $\operatorname{P_{rel}}(\mathrm{X}_{j} | \mathrm{X}_{i}, \mathrm{C})$ from the relation prediction layer and compute the weighted sum of each kind of relation to update the utterance representation, rather than taking  the hard predicted relation by applying an argmax operation over the distribution.

After encoding the constructed discourse structure with  a message passing process, we obtain the discourse relation-augmented utterance representations $\boldsymbol{\mathcal{R}}$, and then utilize the updated utterance representations to revise the syntactic token states with a linear feed-forward network:
\[
    \boldsymbol{z}_{w_{i,n}} = \boldsymbol{\mathrm{W}}_{1} \cdot\left[\boldsymbol{h}_{w_{i,n}}^{k+1} ; \boldsymbol{r}_{i}^{k+1}\right]+\boldsymbol{b}_1,
\]
where $\boldsymbol{h}_{w_{i,n}}^{k+1}$ refers to the token state of $w_{i,n}$ outputted from the $(k+1)$-th layer of SynGCN, $\boldsymbol{r}_{i}^{k+1}$ refers to the state of the corresponding utterance that the token belongs to, outputted from the $(k+1)$-th layer of RelGCN. The operation thus augments syntactic token states $\boldsymbol{\mathcal{H}}_\mathrm{X}$ and $\boldsymbol{\mathcal{H}}_\mathrm{C}$ with discourse information-based utterance representation to obtain discourse context-based token states $\boldsymbol{\mathcal{Z}}_\mathrm{X}=(\boldsymbol{z}_{w_{1}},\dots,\boldsymbol{z}_{w_{n}})$ and $\boldsymbol{\mathcal{Z}}_\mathrm{C}=(\boldsymbol{z}_{w_{1,i}},\dots,\boldsymbol{z}_{w_{i,{l_i}}})$, which will be used to model the referent semantics of the dropped pronoun in the dropper pronoun recovery layer.

\subsection{Pronoun Recovery Layer}
This layer takes in the revised token representations $\boldsymbol{\mathcal{Z}}_\mathrm{X}$ and $\boldsymbol{\mathcal{Z}}_\mathrm{C}$, and attempts to find tokens in context $\mathrm{C}$ that describe the referent of the dropped pronoun in the pro-drop utterance $\mathrm{X}$ with an attention mechanism. The referent representation is then captured as the weighted sum of discourse context-based token states as:
\vspace{-6pt}
\[aw_{i,i',n'}=\operatorname{softmax}(\boldsymbol{\mathrm{W}}_2\left(\boldsymbol{z}_{w_{i}} \odot \boldsymbol{z}_{w_{i',n'}}\right) \\ +b_2),\]
\vspace{-12pt}
\[\boldsymbol{r}_{w_i}=\sum_{i'=1}^{m}\sum_{n'=1}^{l_{i'}} aw_{i, i', n'} \cdot \boldsymbol{z}_{w_{i',n'}}.\]
\vspace{-6pt}

Then we concatenate the referent representation $\boldsymbol{r}_{w_i}$ with the syntactic token representation $\boldsymbol{h}_{w_i}^{k+1}$ to predict the dropped pronoun category as follows: 
\[\boldsymbol{hr}_{w_i}=\tanh \left(\boldsymbol{\mathrm{W}}_3 \cdot\left[\boldsymbol{h}_{w_i}^{k+1} ; \boldsymbol{r}_{w_i}\right]+\boldsymbol{b}_3\right), \]
\[P\left(y_{i} | w_{i}, C\right)=\operatorname{softmax}\left(\boldsymbol{\mathrm{W}}_4 \cdot \boldsymbol{hr}_{w_i}+\boldsymbol{b}_4\right).\]

The objective of dropped pronoun recovery aims to minimize cross-entropy between the predicted label distributions
and the annotated labels for all sentences as:
\begin{equation}
    \operatorname{loss_{dp}}=-\sum_{q \in Q} \sum_{i=1}^{l_i} \delta\left(y_{i}|w_{i}, \mathrm{C}\right) \log \left(P\left(y_{i}|w_{i}, \mathrm{C}\right)\right), \nonumber
\end{equation}
where $Q$ represents all training instances, $l_i$ represents the number of words in pro-drop utterance; $\delta\left(y_{i}|w_{i}, \mathrm{C}\right)$ represents the annotated label of $w_{i}$.

\subsection{Training Objective}
We train our DiscProReco by jointly optimizing the objective of both discourse relation prediction and dropped pronoun recovery. The total training objective is defined as:
\begin{equation}
\label{loss_function}
\operatorname{loss} = \alpha \cdot (\operatorname{loss_{arc}} + \operatorname{loss_{label}}) + \beta \cdot \operatorname{loss_{dp}},
\end{equation}
where $\alpha$ and $\beta$ are weights of CDP objective function and DPR objective function respectively.

\section{The SPDPR Dataset}
\label{dataset_construct}
To verify the effectiveness of DiscProReco, we need a conversational corpus containing the annotation of both dropped pronouns and discourse relations. To our knowledge, there is no such a public available corpus. Therefore, we constructed the first Structure Parsing-enhanced Dropped Pronoun Recovery (SPDPR) dataset by annotating the discourse structure information on a popular dropped pronoun recovery dataset (i.e., Chinese SMS).

The Chinese SMS/Chat dataset consists of 684 multi-party chat files and is a popular benchmark for dropped pronoun recovery~\citep{Yang:15}. In this study, we set the size of the context snippet to be 8 utterances which include the current pro-drop utterance plus 5 utterances before and 2 utterances after. When performing discourse relation annotation we ask three linguistic experts to independently choose a head utterance for the current utterance from its context and annotate the discourse relation between them according to a set of 8 pre-defined relations (see Appendix A). The inter-annotator agreement for discourse relation annotation is 0.8362, as measured by Fleiss's Kappa. The resulting SPDPR dataset consists of 292,455 tokens and 40,280 utterances, averaging 4,949 utterance pairs per relation, with a minimum of 540 pairs for the least frequent relation and a maximum of 12,252 for the most frequent relation. The SPDPR dataset also annotates 31,591 dropped pronouns (except the ``None'' category).

\begin{table*}
\begin{center}
\resizebox{1.00\hsize}{!}{
\begin{tabular}{p{5.6cm}|p{0.83cm}<{\centering} p{0.83cm}<{\centering} p{0.83cm}<{\centering}| p{0.83cm}<{\centering} p{0.83cm}<{\centering} p{0.83cm}<{\centering}| p{0.83cm}<{\centering} p{0.83cm}<{\centering} p{0.83cm}<{\centering}}
 			\toprule[1pt]
 			 & \multicolumn{3}{c|}{\textbf{SPDPR}} & \multicolumn{3}{c|}{\textbf{TC of OntoNotes}} & \multicolumn{3}{c}{\textbf{BaiduZhidao}} \\ \cmidrule(r){2-4}  \cmidrule(r){5-7} \cmidrule(r){8-10}  
			\textbf{Model} &  \textbf{P(\%)} &  \textbf{R(\%)} & \textbf{F(\%)} & \textbf{P(\%)} & \textbf{R(\%)} & \textbf{F(\%)} & \textbf{P(\%)} & \textbf{R(\%)} & \textbf{F(\%)} \\ \midrule[0.5pt]
			MEPR & 37.27 & 45.57 & 38.76 & - & - & - & - & - & - \\ 
			NRM & 37.11 & 44.07 & 39.03 & 23.12 & 26.09 & 22.80 & 26.87 & 49.44 & 34.54 \\ 
			BiGRU & 40.18 & 45.32 & 42.67 & 25.64 & 36.82 & 30.93 & 29.35 & 42.38 & 35.83 \\ 
			NDPR & 49.39 & 44.89 & 46.39 & 39.63 & 43.09 & 39.77 & 41.04 & 46.55 & 42.94 \\
			XLM-RoBERTa-NDPR & 54.03 & 50.18 & 52.46 & 43.14 & 46.37 & 45.13 & 46.04 & 49.12 & 47.54\\
			Transformer-GCRF & 52.51 & 48.12 & 49.81 & 40.48 & 44.64 & 42.45 & 43.30 & 46.54 & 43.92 \\
			\hline
			DiscProReco & 59.58 & 53.68 & 57.37 & - & - & - & - & - & - \\ 
			DiscProReco(XLM-R-w/o RelGCN) & 56.32 & 52.28 & 55.67 & \bf 44.62 &\bf 47.14 &\bf 46.98 &\bf 47.31 &\bf 50.43 &\bf 48.19 \\ 
			DiscProReco(XLM-R) &\bf 61.13 &\bf 54.26 &\bf 59.47 & - & - & - & - & - & - \\
			\bottomrule[1pt]
		\end{tabular}}
	\caption{\label{comparable-result} Experimental results produced by the baseline models, the proposed model DiscProReco and two variants of DiscProReco on all three conversation datasets in terms of precision, recall and F-score.}
 	\end{center}
\end{table*}

\section{Experiments}
\label{experiment}
\subsection{Experimental Settings}
In this work, 300-dimensional pre-trained embeddings~\citep{DBLP:conf/acl/LiZHLLD18} were input to the BiGRU encoder, and 500-dimensional hidden states were uitilized. For SynGCN and RelGCN, we set the number of GCN layers as 1 and 3 respectively, and augment them with a dropout rate of 0.5. The Stanza dependency parser~\citep{qi2020stanza} returns 41 kinds of dependency edges. We remove 13 types of them which connects the punctuation with other tokens, and irrelevant to referent description. During training, we utilized Adam optimizer~\citep{kingma2014adam} with a 0.005 learning rate and trained our model for 30 epochs. The model performed best on the validation set is used to make predictions on the test set. We repeat each experiment 10 times and records the average results.


\subsection{Dropped Pronoun Recovery}
\textbf{Datasets and Evaluation Metrics} 
We tested the performance of DiscProReco for DPR on three datasets: (1) TC section of OntoNotes Release 5.0, which is a transcription of Chinese telephone conversations, and is released in the CoNLL 2012 Shared Task. (2) BaiduZhidao, which is a question answering corpus~\citep{zhang:neural}. Ten types of concrete pronouns were annotated according to the pre-defined guidelines. 
These two benchmarks do not contain the discourse structure information and are mainly used to evaluate the effectiveness of our model for DPR task. (3) The SPDPR dataset, which contains 684 conversation files annotated with dropped pronouns and discourse relations. Following practice in~\citep{Yang:15,Yang2019NDPR}, we reserve the same 16.7\% of the training instances as the development set, and a separate test set was used to evaluate the models. The statistics of the three datasets are shown in Appendix B. 

Same as existing efforts~\citep{Yang:15,Yang2019NDPR}, we use Precision(P), Recall(R) and F-score(F) as metrics when evaluating the performance of dropped pronoun models. 

\noindent\textbf{Baselines}
We compared DiscProReco against existing baselines, including:
(1) MEPR~\citep{Yang:15}, which leverages 
a Maximum Entropy classifier to predict the type of  dropped pronoun before each token;
(2) NRM~\citep{zhang:neural}, which employs two MLPs to predict the position and  type of a dropped pronoun separately; 
(3) BiGRU, which utilizes a bidirectional GRU to encode each token in a pro-drop sentence and then makes prediction;
(4) NDPR~\citep{Yang2019NDPR}, which models the referents of dropped pronouns from a large context with a structured attention mechanism;
(5) Transformer-GCRF~\citep{Yang2020}, which jointly recovers the dropped pronouns in a conversational snippet with general conditional random fields; (6) XLM-RoBERTa-NDPR, which utilizes the pre-trained multilingual masked language model~\citep{DBLP:conf/acl/ConneauKGCWGGOZ20} to encode the pro-drop utterance and its context, and then employs the attention mechanism in NDPR to model the referent semantics.

We also compare two variants of DiscProReco:
(1) DiscProReco (XLM-R-w/o RelGCN), which replaces the BiGRU encoder with the pre-trained XLM-RoBERTa model, removes the RelGCN layer, and only utilizes SynGCN to encode syntactic token representations for predicting the dropped pronouns.
(2) DiscProReco(XLM-R) which uses the pre-trained XLM-RoBERTa model as an encoder to replace the BiGRU network in our proposed model.

\noindent\textbf{Experimental Results}
Table~\ref{comparable-result} reports the results of DiscProReco and the baseline methods on DPR. Please note that for the baseline methods, we directly used the numbers originally reported in the corresponding papers. From the results, we observed that our variant model DiscProReco(XLM-R-w/o RelGCN) outperforms existing baselines on three datasets by all evaluation metrics, which prove the effectiveness of our system as a stand-alone model for recovering  dropped pronouns. 
We attribute this to the ability of our model to consider long-distance syntactic dependencies between tokens in the same utterance. Note that the results for feature-based baseline MEPR~\citep{Yang:15} on OntoNotes, and BaiduZhidao are not available because several essential features cannot been obtained. However, our proposed DiscProReco still significantly outperforms DiscProReco (XLM-R-w/o RelGCN) as it achieved $3.26\%$, $1.40\%$, and $1.70\%$ absolute improvements in terms of precision, recall and F-score respectively on SPDPR corpus. This shows that discourse relations between utterances are crucially important for modeling the referent of dropped pronouns and achieving better performance in dropped pronoun recovery. This is consistent  with the observation in~\citep{ghosal2019dialoguegcn}. The best results are achieved when  our model uses uses the pre-trained XLM-RoBERTa (i.e., DiscProReco(XLM-R)). Note that discourse relations are not available for Ontonotes and BaiduZhidao datasets and thus we do not have joint learning results for these two data sets.

\begin{figure}
	\centering
	\includegraphics[width=0.46\textwidth]{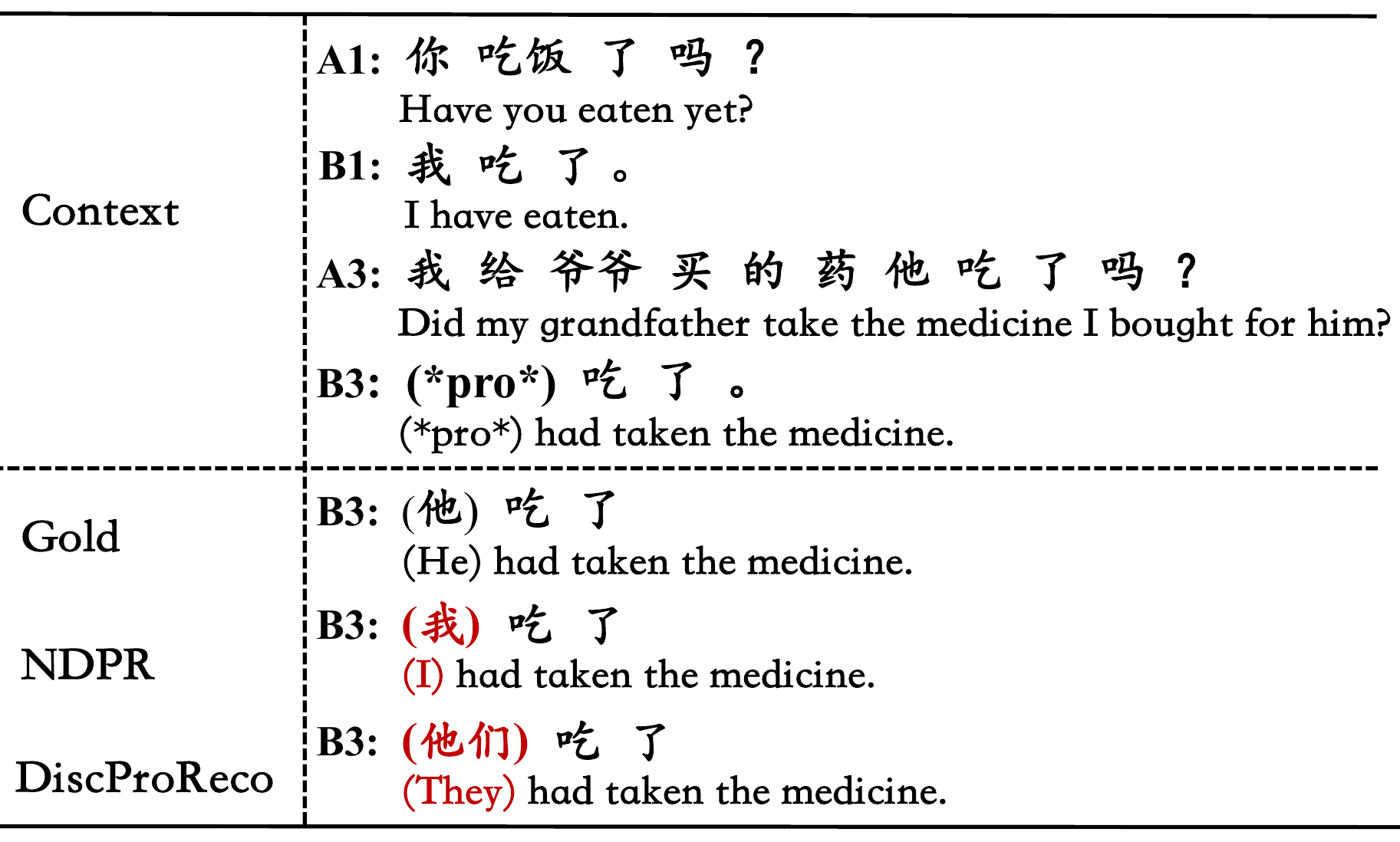}
	\caption{Results of different DPR models.}
	\label{error_case}
\end{figure}

\noindent\textbf{Error Analysis}
We further investigated some typical mistakes made by our DiscProReco for DPR. Resolving DPR involves effectively modeling the referent of each dropped pronoun from the context to recover the dropped pronoun. As illustrate in Figure~\ref{error_case}, both DiscProReco and NDPR model the referent from the context. The former outperforms the latter since it considers the conversation structure that the utterance B3 is a reply to A3 but not an expansion to the utterance B1. However, just modeling the referent from the context is insufficient. In Figure~\ref{error_case}, the referent of the dropped pronoun was correctly identified but the dropped pronoun is mistakenly identified as ``(他们/they)''. This indicates that the model needs to be augmented with some additional knowledge, such as the difference between singular and plural pronouns.

\subsection{Conversational Discourse Parsing}
\noindent\textbf{Datasets and Evaluation Metrics}
We evaluated the effectiveness of our DiscProReco framework for CDP task on two datasets as: (1) STAC, which is a standard benchmark for discourse parsing on multi-party dialogue~\citep{8fb5af5c57eb43d7afaa5f7b5c95ad2a}. The dataset contains 1,173 dialogues, 12,867 EDUs and 12,476 relations. Same as existing studies, we set aside 10\% of the training dialogues as the validation data. (2) SPDPR, which is constructed in our work containing 684 dialogues and 39,596 annotated relations. Following~\citep{shi2018a}, we also utilized micro-averaged F-score as the evaluation metric. 

\begin{table}[t!]
\begin{center}
\resizebox{1.0\hsize}{!}{
\begin{tabular}{p{3.7cm}|p{0.6cm}<{\centering}p{1.2cm}<{\centering}|p{0.7cm}<{\centering}p{1.2cm}<{\centering}}
\toprule[1pt]
 & \multicolumn{2}{c|}{\textbf{STAC}} & \multicolumn{2}{c}{\textbf{SPDPR}} \\ \cmidrule(r){2-3}  \cmidrule(r){4-5}
\textbf{Model} & \textbf{Arc} & \textbf{Rel} & \textbf{Arc} & \textbf{Rel} \\ \hline 
MST & 68.8 & 50.4 & - & - \\
ILP & 68.6 & 52.1 & - & - \\
Deep+MST & 69.6 & 52.1 & 81.06 & 40.93 \\
Deep+ILP & 69.0 & 53.1 & 80.53 & 41.38 \\
Deep+Greedy & 69.3 & 51.9 & 81.32 & 42.38 \\
Deep Sequential & 73.2 & 55.7 & 83.00 & 43.45 \\ \hline
DiscProReco(w/o DPR) &\bf 74.1 &\bf 57.0 & 84.51 & 51.34 \\ 
DiscProReco & - & - &\bf 87.97 &\bf 53.07 \\ \bottomrule[1pt]
\end{tabular}}
\caption{\label{logical_relation_result} Micro-averaged F-score (\%) of conversational discourse parsing on two standard benchmarks.}
\end{center}
\end{table}

\noindent\textbf{Baselines}
We compared our DiscProReco with existing baseline methods:
(1) MST~\citep{afantenos2015discourse}: A approach that uses local information in two utterances to predict the discourse relation, and uses the Maximum Spanning Tree (MST) to construct the discourse structure;
(2) ILP~\citep{perret2016integer}: Same as MST except that the MST algorithm is replaced with Integer Linear Programming (ILP); 
(3) Deep+MST: A neural network that encodes the discourse representations with GRU, and then uses MST to construct the discourse structure;
(4) Deep+ILP: Same as Deep+MST except that the MST algorithm is replaced with Integer Linear Programming (ILP); 
(5) Deep+Greedy: Similar to Deep+MST and Deep+ILP except that this model uses a greedy decoding algorithm to select the parent for each utterance;
(6) Deep Sequential~\citep{shi2018a}: A deep sequential neural network which predicts the discourse relation utilizing both local and global context.

In order to explore the effectiveness of joint learning scheme, we also make a comparison of our DiscProReco with its variant, referred to as DiscProReco(w/o DPR), which predict the discourse relation independently, without recovering the dropped pronouns.     

\noindent\textbf{Experimental Results}
We list the experimental results of our approach and the  baselines in Table~\ref{logical_relation_result}. For the STAC dataset, we also reported the original results of the STAC benchmark from an existing paper~\citep{shi2018a}, and apply our DiscProReco to this corpus. For the SPDPR dataset, we ran the baseline methods with the same parameter settings. From the results we can see that the variant of our approach DiscProReco (w/o DPR) outperforms the baselines of discourse parsing. We attribute this to the effectiveness of the biaffine attention mechanism for dependency parsing task~\citep{2020A,ji2019graph-based}. However, our approach DiscProReco still significantly outperforms all the compared models. We attribute this to the joint training of the CDP task and the DPR task. The parameter sharing mechanism makes these two tasks benefits each other. Note that the results for the joint model is not available for STAC as STAC is not annotated with dropped pronouns.

\subsection{Interaction between DPR and CDP}
We also conducted experiments on SPDPR to study the quantitative interaction between DPR and CDP.  
Firstly, during the training process, we optimize our DiscProReco model utilizing the objective function in Eq.~\ref{loss_function} until the CDP task achieves a specific F-score (i.e., gradually increases from 30.64 to 50.38). Then we fix the CDP components and continue to optimize the components of DPR task. We conduct this experiment to explore the influence of CDP task on the DPR task. Secondly, we set the ratio between $\alpha$ and $\beta$ in Eq.~\ref{loss_function} varies from 0.25 to 1.25 and record the F-score of DPR and CDP respectively. We conduct this experiment to study the interanction between these two tasks by modifying their weights in the objective function.

\begin{figure}[t]
    \centering
\begin{minipage}[b]{0.4\textwidth}
\includegraphics[width=1\textwidth]{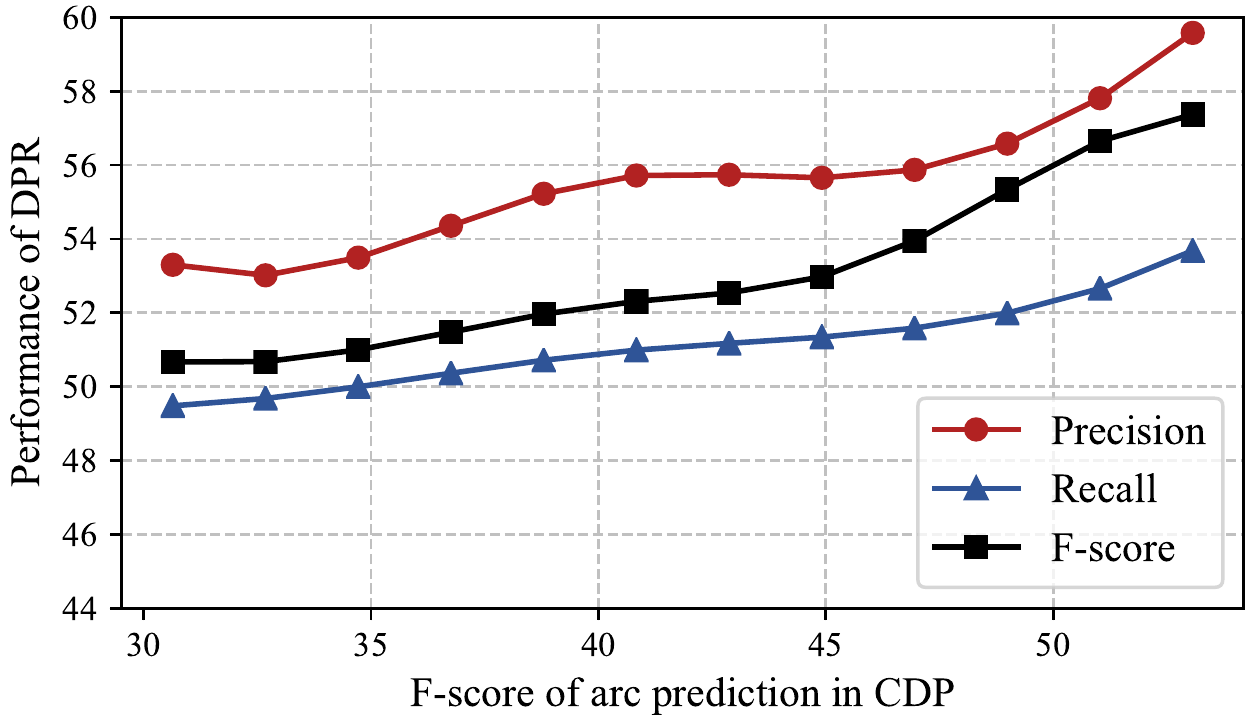} \\
\includegraphics[width=1\textwidth]{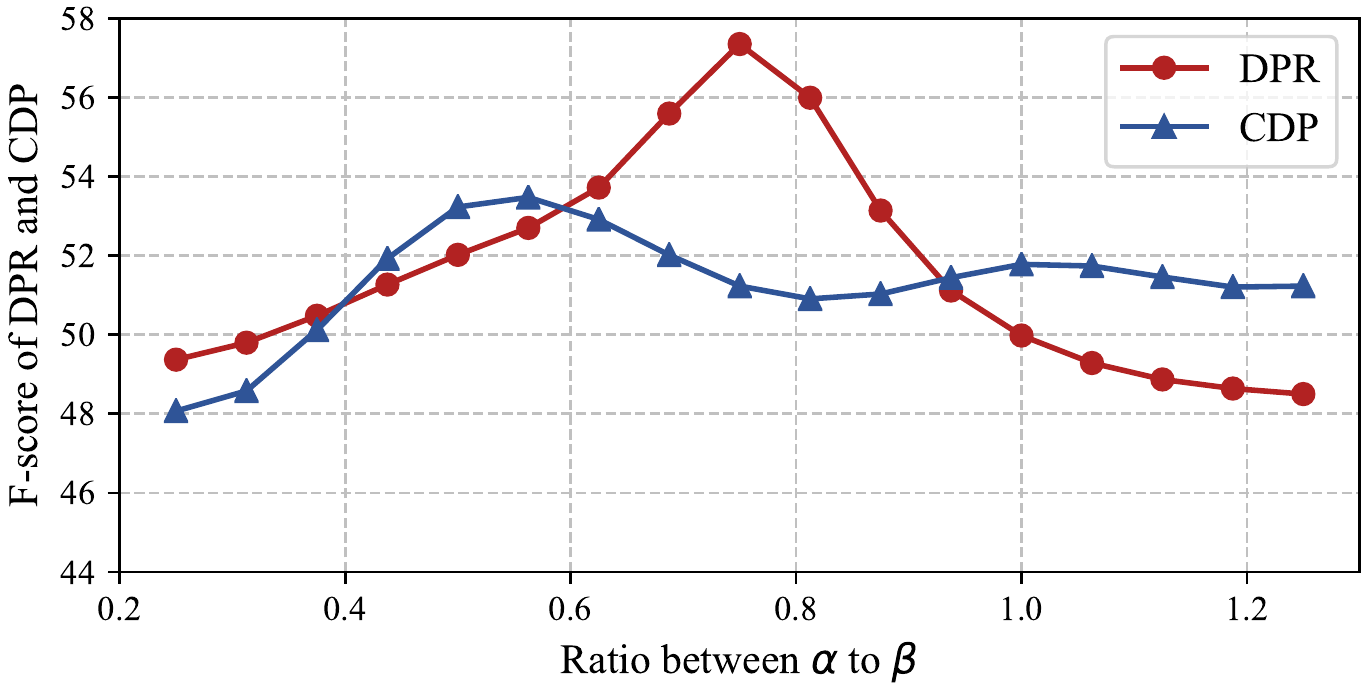}
\end{minipage}
    \caption{Exploratory results. (a) Interaction between DPR and CDP; (b) Effects of parameters (i.e., $\alpha$ and $\beta$)}
    \label{exploratory}
\end{figure}

Results of these two experiments are shown in Figure~\ref{exploratory}. 
According to Figure~\ref{exploratory} (a), the performance of DPR is increased in terms of all evaluation metrics as the F-score of CDP increases, which indicates that exploring the discourse relations between utterances benefits dropped pronoun recovery. Moreover, Figure~\ref{exploratory} (b) illustrate the performance of DPR and CDP 
when the ratio between $\alpha$ to $\beta$ varies gradually. Results show that the performance of CDP remains stable, while the performance of DPR increases at beginning and then decrease sharply as the ratio increases, indicating that DiscProReco framework should pay more attention to DPR during the optimizing process.

\section{Related Work}
Dropped pronoun recovery is a critical technique that can benefit many downstream applications~\citep{wang:2016a, wang:2018, su2019improving}.~\citet{Yang:15} for the first time proposed this task, and utilized a Maximum Entropy classifier to recover the dropped pronouns in text messages.
\citet{giannella:17} further employed a linear-chain CRF to jointly predict the position and type of the dropped pronouns in a single utterance using hand-crafted features. 
Due to the powerful semantic modeling capability of deep learning, \citet{zhang:neural,Yang2019NDPR} introduced neural network methods to recover the dropped pronoun by modeling its semantics from the context. 
All these methods represent the utterances without considering the relationship between utterances, which is important to identify the referents. 
Zero pronoun resolution is also a closely related line of research to DPR~\citep{Chen:16,Yin:17,yin:2018}. The main difference between DPR and zero pronoun resolution task is that DPR considers both anaphoric and non-anaphoric pronouns, and doesn't attempt to resolve it to a referent.

Existing discourse parsing methods first predicted the probability of discourse relation, and then applied a decoding algorithm to construct the discourse structure~\citep{Muller2012Constrained, li2014recursive, afantenos2015discourse,perret2016integer}. 
A deep sequential model~\citep{shi2018a} was further presented to predict the discourse dependencies utilizing both local information of two utterances and the global information of existing constructed discourse structure. All these methods consider how to do relation prediction independently. However, in this work, we explore the connection between the CDP and DPR, and attempt to make these two tasks mutually enhance each other.

\section{Conclusion}
This paper presents that dropped pronoun recovery and conversational discourse parsing are two strongly related tasks. To make them benefit from each other, we devise a novel framework called DiscProReco to tackle these two tasks simultaneously. The framework is trained in a joint learning paradigm, and the parameters for the two tasks are jointly optimized. To facilitate the study of the problem, we created a large-scale dataset called SPDPR which contains the annotations of both dropped pronouns and discourse relations. Experimental results demonstrated that DiscProReco outperformed all baselines on both tasks. 

\section*{Acknowledgments}
This work was supported by the National Key R\&D Program of China (2019YFE0198200), the National Natural Science Foundation of China (No. 61872338, No. 61832017), Beijing Academy of Artificial Intelligence (BAAI2019ZD0305), Beijing Outstanding Young Scientist Program NO. BJJWZYJH012019100020098 and BUPT Excellent Ph.D. Students Foundation (No.CX2020305).

\bibliographystyle{acl_natbib}
\bibliography{acl2021}

\clearpage

\appendix
\section{Discourse Relations}
The discourse relation describes a participant may speak a utterance to agree with, respond to, or indicate understanding of another utterance in the conversational context. According to~\cite{xue2016annotating} , each utterance is assumed only related to one previous utterance. All relations are summarized as 6 types between \textit{same-participant} utterance pairs, and 2 types between \textit{different-participant} utterance pairs, as summarized in Table~\ref{relation-description}.

\section{Statistics of DPR Datasets}
The statistics of three dropped pronoun recovery benchmarks (i.e., SPDPR, TC section of OntoNotes and BaiduZhidao) are shown in Table~\ref{dataset_statistic}.

\begin{table}
\begin{center}
\begin{tabular}{p{2.5cm}|p{10.7cm}}
\Xhline{1.2pt}
\textbf{Relation} & \textbf{Description} \\ \hline
\multicolumn{2}{c}{Different Participant} \\ \hline
Agreement & a participant provides a response to a previous request or suggestion \\
Understanding & a participant indicates understanding of a previous utterance \\
Directive & a participant asks another one to do something \\
Question & a general request for another participant \\
Answer & a participant provides the information requested by another participant \\
Feedback & a participant responds to another speaker's utterance \\ \hline
\multicolumn{2}{c}{Same Participant} \\ \hline
Expansion & a participant provides an elaboration of a previous utterance \\
Contingency & a participant continues to say something else \\
\Xhline{1.2pt}
\end{tabular}
\caption{\label{relation-description}  Explanation of discourse relations.}
\end{center}
\end{table}

\begin{table}
\begin{center}
\begin{tabular}{p{1.0cm}<{\centering}|p{1.45cm}<{\centering}c|p{1.45cm}<{\centering}c}
\Xhline{1.2pt}
& \multicolumn{2}{c|}{Training} & \multicolumn{2}{c}{Test} \\ \cline{2-5}
& \#Sentences & \#DPs & \#Sentences & \#DPs \\ \hline
SPDPR & 35,933 & 28,052 & 4,346 & 3,539\\
TC & 6,734 & 5,090 & 1,122 & 774\\
Zhidao & 7,970 & 5,097 & 1,406 & 786\\
\Xhline{1.2pt}
\end{tabular}
\end{center}
\caption{\label{dataset_statistic} Statistics of training and test sets on three conversational datasets.}
\end{table}

\end{CJK*}
\end{document}